\documentclass[conference]{IEEEtran}
\IEEEoverridecommandlockouts

\usepackage{cite}
\usepackage{amsmath,amssymb,amsfonts}
\usepackage{algorithmic}
\usepackage{graphicx}
\usepackage{textcomp}
\usepackage{xcolor}
\usepackage{multirow}
\usepackage{tabularx}
\usepackage{subcaption}    
\usepackage[T1]{fontenc}
\usepackage{dblfloatfix} 
\def\BibTeX{{\rm B\kern-.05em{\sc i\kern-.025em b}\kern-.08em
    T\kern-.1667em\lower.7ex\hbox{E}\kern-.125emX}}
\begin{document}

\title{PRIME: Physics-Related Intelligent Mixture of Experts for Transistor Characteristics Prediction

\thanks{
\IEEEauthorrefmark{4} Corresponding author.}
}

\author{\IEEEauthorblockN{Zhenxing Dou\IEEEauthorrefmark{1}, Yijiao Wang\IEEEauthorrefmark{1}\IEEEauthorrefmark{4}, Tao Zou\IEEEauthorrefmark{1}, Zhiwei Chen\IEEEauthorrefmark{1},\\
Fei Liu\IEEEauthorrefmark{3}, Peng Wang\IEEEauthorrefmark{1}\IEEEauthorrefmark{4},  Weisheng Zhao\IEEEauthorrefmark{1}}

\IEEEauthorblockA{\IEEEauthorrefmark{1}School of Integrated Circuit Science and Engineering, Beihang University, Beijing, China}


\IEEEauthorblockA{\IEEEauthorrefmark{3}School of Integrated Circuit, Peking University, Beijing, China \\
Emails: \{zhenxing, YijiaoWang, zoutao41, czw1224, wang.peng, wszhao\}@buaa.edu.cn, \\
feiliu@pku.edu.cn} 
}

\maketitle
\begin{abstract}


In recent years, machine learning has been extensively applied to data prediction during process ramp-up, with a particular focus on transistor characteristics for circuit design and manufacture. However, capturing the nonlinear current response across multiple operating regions remains a challenge for neural networks. To address such challenge, a novel machine learning framework, PRIME (Physics-Related Intelligent Mixture of Experts), is proposed to capture and integrate complex regional characteristics. 
In essence, our framework incorporates physics-based knowledge with data-driven intelligence. 
By leveraging a dynamic weighting mechanism in its gating network, PRIME adaptively activates the suitable expert model based on distinct input data features.  
Extensive evaluations are conducted on various gate-all-around (GAA) structures to examine the effectiveness of PRIME and considerable improvements (60\%-84\%) in prediction accuracy are shown over state-of-the-art models.

\end{abstract}


\section{Introduction}

As transistor sizes continue to shrink, physical, technological, and economic limitations have been continuously pushing Moore's Law to its limits\cite{leiserson2020there}. To sustain technological advancements as device scaling approaches these limits, there is active research on emerging devices\cite{kim2024future}. Technology Computer-Aided Design (TCAD) employs the governing laws on device physics and provides precise simulation on its electrical performance \cite{bhoj2012efficient, ota2017perspective, sathaiya2022comprehensive}. 
But it often requires substantial computational resources for decent accuracy. To address this challenge, compact models, such as BSIM \cite{chauhan2012bsim}, HiSim \cite{mattausch2014hisim} and UTSOI \cite{pan2007novel}), are extracted from TCAD simulations and characterize the electrical properties of transistors\cite{myung2024new} in physics-based formulations. 

However, due to one's limited understanding of the full  physical properties of new devices, the time-consuming nature of TCAD simulations, and the labor-intensive processes of modeling and parameter extraction, those physical models are struggling to keep pace with rapid technological advancements. In recent years, data-driven machine learning (ML), has emerged as a promising approach in semiconductor manufacturing and simulation\cite{wong2020tcad, raju2020application, sheelvardhan2023machine}. For example, Wang et al.\cite{wang2021artificial} established an artificial neural network (ANN) architecture for field-effect transistors (FETs). 
By carefully selecting conversion functions and loss functions for ANN training, their predictive model can reproduce the current-voltage ($I-V$) characteristics of advanced FETs with decent accuracy. Meanwhile, Yang et al. \cite{yang2021transistor} employed a multigradient neural network (MNN) to capture both the direct current and alternating current characteristics of transistors, as well as their high-order derivatives, making it applicable to various device types. Alternatively, Mehta and Wong \cite{mehta2020prediction} proposed utilizing an autoencoder to extract latent features associated with physical parameters and constructed third degree polynomial regression to predict the full transistor $I-V$ curves. Following this approach, Wang et al. \cite{wang2023drpce} recently proposed the differential residual polynomial chaos expansion network (DrPCE-Net). It first utilizes polynomial chaos expansion for prior estimation and then reduces its error through residual connections, achieving good fitting results on gate-all-around (GAA) devices.

Despite the success of data-driven neural network methods, they still struggle with generalization across different types of devices and problems. Constructed as black-box models, neural networks focus on data fitting while neglecting the physical characteristics of devices. As a result, such a network is designated for very specific problem. In other words, network suitable for one characteristics region may become problematic in other operating regions.

To address such challenge, we propose Physics-Related Intelligent Mixture of Experts (PRIME), a novel approach for predicting transistor characteristics. Based on the concept of Mixture of Experts, PRIME incorporates physics expertise on transistor operating regions and provides accurate and integrated predictions. Its novelty can be summarized as below: 
\begin{itemize}
    \item \textbf{Novel Model Framework.} \\
    To the best of our knowledge, this is the first neural network architecture that utilizes physics information for its characteristics prediction.  

    \item \textbf{Integrated Prediction.} \\
    By incorporating expert information to simulate distinct regions of the transistor, PRIME provides an integrated prediction on transistor characteristics and ensures robustness across various device geometries.

    \item \textbf{Significant Accuracy Improvement.} \\
    Comparing to state-of-the-art methods, PRIME achieves  60\% to 84\% improvement in predictive accuracy for GAA transistor characteristics.



\end{itemize}

The remaining sections of this paper are organized as follows: In Section \ref{section 2}, we formulate the general problem and introduce the GAA device and the selected physical parameters. Section \ref{section 3} outlines our Physics-Related Intelligent Mixture of Experts (PRIME) of transistor characteristics prediction. 
Section \ref{section 4} illustrates and discusses the experimental results of our approach along with those of the baseline methods, and conclusions are drawn in Section \ref{section 5}.


\section{PROBLEM DEFINITION AND DEVICE INTRODUCTION}\label{section 2}


\subsection{Problem Definition}

The main characteristics of a transistor are typically described by its $I-V$ curves, which reveal the device's current response under various voltage conditions. We formulate the problem of modeling the $I-V$ curve as a regression task, expressed as $y=f(\boldsymbol{x})$, to capture the intricate relationship between current, voltage and physical parameters. Our goal is to use dataset from commercial technology computer-aided design (TCAD) simulation of devices $D = \{y_i, \boldsymbol{x}^{(i)}\}_{i=1}^{N}$ to approximate the mapping $f(\cdot)$, where $N$ is the size of dataset.

In this problem, the input vector $\boldsymbol{x}=[z_1, z_2, \ldots, z_k, V_{\mathrm {gs}}, V_{\mathrm {ds}}]^{T}$ consists of the physical parameters of the device as well as the applied gate-to-source voltage $V_{\mathrm {gs}}$ and drain-to-source voltage $V_{\mathrm {ds}}$. 
The output is the common logarithm (base 10) of the drain-to-source current $I_{\mathrm {ds}}$, represented as $y = \log_{10}(I_{\mathrm {ds}})$. This logarithmic transformation effectively compresses the dynamic range of $I_{\mathrm {ds}}$, which is particularly useful in the subthreshold region where the current values can vary by several orders of magnitude. Since we predict the logarithmic value of $I_{\mathrm {ds}}$, the resulting $I-V$ curve is based on the condition that $V_{\mathrm {ds}}$ is positive, ensuring that $I_{\mathrm {ds}}$ remains positive. Additionally, we incorporate a physically consistent assumption: when $V_{\mathrm {ds}} = 0$, $I_{\mathrm {ds}}$ is also assumed to be 0, ensuring that the model accurately reflects the actual device behavior. To mitigate the impact of varying physical parameter scales on model training, we implemented the following normalization techniques:
 
 \begin{equation}
 \label{eq1: preprocessed equtation}
    \boldsymbol{x}^{\prime} = \dfrac{\boldsymbol{x} - \boldsymbol{x}_{\min}}{\boldsymbol{x}_{\max} - \boldsymbol{x}_{\min}},
\end{equation}
where $\boldsymbol{x}_{\min}$ represents the minimum value of the physical parameter, and $\boldsymbol{x}_{\max}$ represents its maximum value.

\subsection{Device Introduction}

In this study, we select the GAA device for its excellent gate control capability, which is highly favored by the industry. GAA is also the mainstream device structure at the current sub-5-nanometer technology node. Without loss of generality, it is noted that the method introduced in this paper is applicable to any device and is not limited to GAA device. Physical parameters include the channel length ($l_g$), nanowire radius (for circular and triangular GAAs, $r$), nanowire height/width (for rectangular GAAs, $h/w$), oxide thickness ($t_{ox}$), doping concentration of source and drain ($N_{sd}$), and the relative dielectric constant of oxide layer ($\varepsilon_{ox}$). 

\begin{figure}[!t]
\centering
\includegraphics[width=\columnwidth]{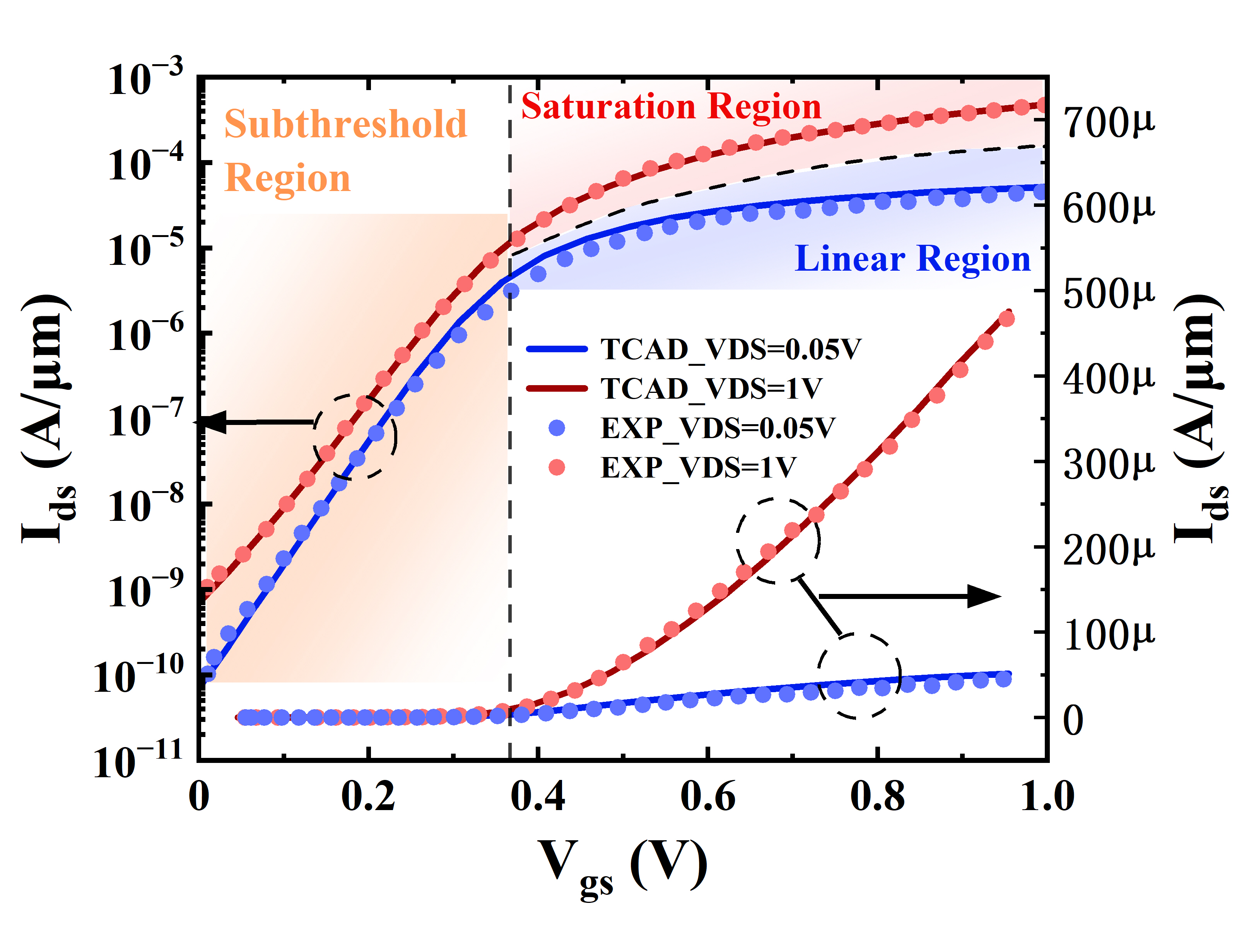}
\caption{Calibration of the transfer characteristics ($I_{\mathrm ds}-V_{\mathrm gs}$) of the GAA nanowire transistor with the experimental data.}
\label{fig1: Calibration}
\end{figure}

\begin{figure*}[!ht]
\centering
\begin{subfigure}{0.28\textwidth}
    \centering
    \includegraphics[width=\textwidth]{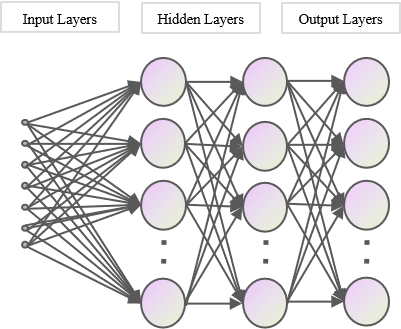}
    \caption{MLP.}
    \label{fig: MLP}
\end{subfigure}
\hspace{0.02\textwidth}
\begin{subfigure}{0.32\textwidth}
    \centering
    \includegraphics[width=\textwidth]{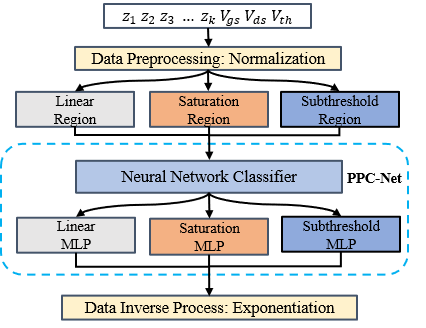}
    \caption{PPC-Net.}
    \label{fig: PPC-Net}
\end{subfigure}
\hspace{0.02\textwidth}
\begin{subfigure}{0.32\textwidth}
    \centering
    \includegraphics[width=\textwidth]{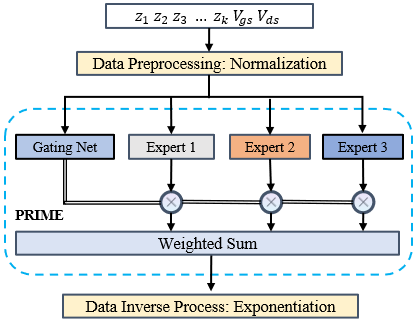}
    \caption{PRIME.}
    \label{fig:PRIME}
\end{subfigure}
\caption{Comparison of MLP, PPC-Net and PRIME. In Figures (a)-(c), black arrows indicate data flow. In Figure (c),the multiplication sign in the circle represents multiplication, and the double solid lines represent the gating network output weights.}
\label{fig:three_models}
\end{figure*}

Figure~\ref{fig1: Calibration} shows the calibration results of the GAA model with the experimental results, which are represented in the form of transfer characteristics \( I_{\mathrm {ds}} - V_{\mathrm {gs}} \) curves for \( V_{\mathrm {ds}} = 0.05\,\text{V}, 0.7\,\text{V} \). The plot highlights three operating regions of the GAA transistor: the subthreshold region (bottom left), linear region (middle), and saturation region (top right). The vertical dashed line indicates the threshold voltage $V_{\mathrm {th}}$, serving as a boundary between regions, while the diagonal dashed line represents the $I_{\mathrm{ds}}$ values at different $V_{\mathrm {ds}} $ levels on each side. The parameters of the GAA model for calibration are listed below: circular cross-sectional shape, \( l_{g} = 14\,\text{nm} \), \( r = 3\,\text{nm} \), \(  t_{ox} = 1\,\text{nm} \), \(  N_{sd} = 1\,\times 10^{20}\text{cm}^{-3} \), \(  \varepsilon_{ox}= 3.9\).

As shown in Fig.~\ref{fig1: Calibration}, for GAA devices, when the gate voltage is less than the threshold voltage $V_{\mathrm {th}}$, the transistor operates in the subthreshold region, where the drain current is not zero and exhibits exponential growth characteristics. The device turns on when $V_{\mathrm {gs}} > V_{\mathrm {th}}$. Depending on the drain-source voltage, the device can be further divided into the linear region and the saturation region. The $I_{\mathrm {ds}}$ formulations in the different operating regions of a transistor are fundamentally distinct, making it a challenge to establish a single mapping function $f(\cdot)$ that can universally describe the current across all three regions simultaneously. 





\section{PROPOSED PRIME METHOD}\label{section 3}

\subsection{Physical Property Classification Neural Network}

A natural approach is to pre-divide the dataset into subthreshold, linear, and saturation regions based on the relative magnitudes of $V_{\mathrm {ds}}$, $V_{\mathrm {gs}}$ and $V_{\mathrm {th}}$. Three independent Multi-Layer Perceptrons (MLP, as shown in Fig.~\ref{fig: MLP}) are trained accordingly, each is dedicated to predict the current for a specific region. To ensure accurate predictions, an additional three-way classifier \cite{memisevic2010gated} must be constructed. The role of this classifier is to analyze the features of the input data $\boldsymbol{x}$ and determine which region it belongs to: subthreshold, linear or saturation. For example, if the classifier predicts that $\boldsymbol{x}$ is in the linear region, the data is passed to the Linear MLP. This allows the model to route the data to the appropriate regional model based on the classification result, thus enhancing the accuracy of the prediction. We refer to this structure as a Physical Property Classification Neural Network (PPC-Net) because it classifies data based on physical properties, like operating regions, to select a specialized model for each region, and the overall framework is shown in Fig.~\ref{fig: PPC-Net}. This structure ensures specialized treatment of each region's characteristics, improving prediction accuracy, but it also has two main drawbacks: 1) The need for pre-classification of data $\boldsymbol{x}$ brings extra overhead. 2) Multiple models must be trained independently, leading to longer training times.

\subsection{Physics-Related Intelligent Mixture of Experts }

Inspired by the mixture of experts (MoE) models in large language models (LLMs), we have incorporated the automatic expert selection mechanism from LLMs into PPC-Net \cite{zhou2022mixture}. Specifically, compared to Fig.~\ref{fig: PPC-Net}, Fig.~\ref{fig:PRIME} retains the three expert models based on physical knowledge. An adaptive gating network replaces the classifier in PPC-Net, dynamically weighting the expert models based on input data. This modification allows the model to automatically learn and apply the characteristics of different regions, effectively calling upon the appropriate expert models. This improvement not only removes the need for pre-classifying data but also effectively prevents regional misclassification issues in PPC-Net, significantly enhancing the accuracy and robustness of transistor characteristics predictions. Based on these advantages, we refer to the improved model as the Physics-Related Intelligent Mixture of Experts (PRIME).

In PRIME, we introduce the Softmax function for the gating network. The Softmax function \cite{bishop2006pattern} is a mathematical function that converts a vector of real values into a probability distribution. For a given input vector $\boldsymbol{x} = [x_1, x_2, \dots, x_k]$, the Softmax function outputs a vector of probabilities $\boldsymbol{p} = [p_1, p_2, \dots, p_k]$ such that each $p_i$ represents the probability associated with class $i$, and the sum of all probabilities equals 1. The formula for the Softmax function is given by:
\begin{equation}
p_j = \frac{\exp(x_j)}{\sum\limits_{j=1}^{k} \exp(x_j)}, \quad  \ j = 1, 2, \dots, k,
\end{equation}
where $\exp(x_j)$ represents the exponential of the input $x_j$, and the denominator ensures that the sum of all output probabilities is normalized to 1. In this model, $k=3$, corresponds to the three experts models.

Fig.~\ref{fig:PRIME} illustrates the workflow of PRIME. First, the raw data  $\boldsymbol{x}$ is normalized according to \eqref{eq1: preprocessed equtation} to prevent discrepancies in magnitude from affecting the model training. After normalization, the data $\boldsymbol{x}^{\prime}$ is processed through three expert models. Each is a separate MLP consisting of $N_{l}$ hidden layers. Its $l$-th hidden layer is denoted as $\boldsymbol{x}_l$, ($l=1,2, \dots  N_{l}$):
\begin{equation}
\begin{aligned}
\boldsymbol{x}_l & =\phi\left(\boldsymbol{w}_{l-1} \boldsymbol{x}_{l-1}+\boldsymbol{b}_{l-1}\right), \\ 
y_{e}^{j} &= \boldsymbol{w}_{N_l} \boldsymbol{x}_{N_l}+\boldsymbol{b}_{N_l}.
\end{aligned}
\end{equation}
Here $\boldsymbol{w}_l$ are the weight matrices and $\boldsymbol{b}_l$ are the bias vectors, $y_{e}^{j}$ represents the output prediction provided by the $j$-th expert model, $j= 1, 2, 3$. Nonlinear activation function $\phi(\cdot)$ is taken as Tanh function \cite{lecun1989handwritten} in each expert model.

The gating network is a critical component of PRIME, responsible for selective weighting, allowing the model to learn regional characteristics autonomously. The gating network assigns dynamic weights to each expert model based on the normalized input $\boldsymbol{x}^{\prime}$, determining their contribution to the final prediction. The input $\boldsymbol{x}^{\prime}$ first passes through an MLP, which generates a set of outputs corresponding to the number of experts in the model. These outputs are then transformed into a set of probability-like weights through the Softmax function, ensuring that the sum of the contribution weights for all experts equals 1. This process can be represented as:
\begin{equation} 
\boldsymbol{p} = \mathrm{Softmax}\left(\text{MLP}(\boldsymbol{x})\right), 
\end{equation}
where $\boldsymbol{p}=[p_{1}, p_{2}, p_{3}]$ represents the weights obtained through the Softmax function. The weight $p_j$ corresponds to the $j$-th expert model, indicating the relative importance of this expert model for the current input. PRIME calculates the final prediction by weighting the outputs of each expert model and summing their contributions: $\tilde{y} = \sum_{j=1}^{3} p_j y_{e}^{j}.$

Through the dynamic weighting mechanism of the gating network, PRIME can adaptively allow different expert models to focus on different aspects of the data based on the characteristics of the input $\boldsymbol{x}$. This enables the model to fully exploit the potential of each part of the input data, maximizing its contribution to model training and prediction accuracy. While improving prediction accuracy, PRIME also effectively reduces redundant computations, enhancing the overall data utilization efficiency of the model.

\subsection{Loss Function}

Earlier research \cite{yang2021transistor} demonstrates that incorporating high-order gradient errors as regularization terms in the loss function helps accelerate the convergence of neural networks and prevent overfitting. To enhance training efficiency and fully utilize the available data, we introduced high-order derivative errors into the training of all models, adding first and second order derivatives to the loss function. These derivatives are computed using finite difference methods \cite{thomas2013numerical}, and the resulting loss function is expressed as follows:
\begin{equation}
\label{loss function}
\begin{aligned}
    \text{loss} &= \dfrac{1}{n} \sum_{i=1}^{n} \left[  (y_{i} - \tilde{y}_{i})^2 + \sum_{m=1}^{2} a_{m} \left( \dfrac{\partial^m y_{i}}{\partial V_{ds_{i}}^{m}} - \dfrac{\partial^m \tilde{y}_{i}}{\partial V_{ds_{i}}^{m}} \right)^2 \right. \\
    & \quad + \left. \sum_{m=1}^{2} b_{m} \left( \dfrac{\partial^m y_{i}}{\partial V_{gs_{i}}^{m}} - \dfrac{\partial^m \tilde{y}_{i}}{\partial V_{gs_{i}}^{m}} \right)^2 \right].
\end{aligned}
\end{equation}

Here, $y_{i}$ represents the target value to be fitted, and $\tilde{y}_{i}$ is the output of the PRIME. $n$ denotes the number of training samples, while $a_{m}$ and $b_{m}$ are the hyperparameters that control the influence of the high-order derivative terms with respect to $V_{\mathrm ds}$ and $V_{\mathrm gs}$ 
on the loss function. We recommend selecting $a_{m}$ and $b_{m}$ within the range of $10^{-3}$ to $10^{-4}$. Larger hyperparameters may lead to underfitting, whereas smaller ones could cause overfitting.

\section{RESULTS AND DISCUSSION}\label{section 4}

\subsection{Experimental Setup}

\begin{figure}[ht]
\centering
\includegraphics[width=0.75\columnwidth]{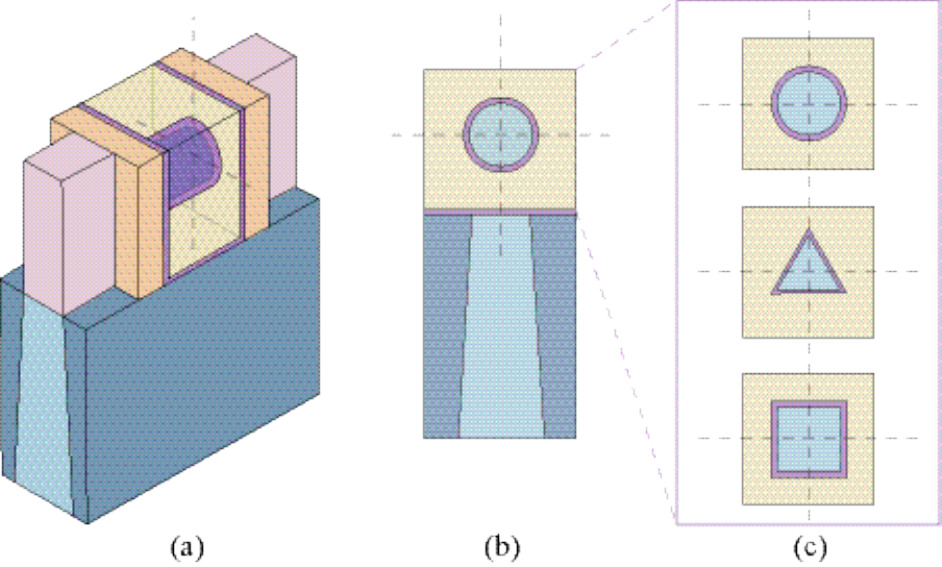}
\caption{GAA transistor structure and three types of cross-section view of GAA transistors. (a) Entire 3D schematic of the device structure. (b) Cross section of entire transistor. (c) Three types of cross sectional shapes: circular, triangular, and rectangular. }
\label{fig3: Nanowire}
\end{figure}

The GAA model is shown in Fig. \ref{fig3: Nanowire} and supports various configurations. To evaluate the model’s generalization across different cross-sectional shapes, we selected three geometries: circular, rectangular and triangular. Based on the physical parameters listed in Table \ref{table 1 Device Parameters Used in the TCAD Simulation}, along with values for $V_{\mathrm {ds}}$ and $V_{\mathrm {gs}}$, we use a commercial TCAD tool to generate a dataset of $I-V$ curves for each device, which is later used for model training and testing. The complete dataset comprises 648 distinct device types and each type contains 225 data points for $I_{\mathrm {ds}}$. For each shape, 400 device types were used for training, and 248 were used for testing. Additionally, we generated 210 discretized $I-V$ data points per device type to allow for error analysis against TCAD simulation data. The threshold voltage $V_{\mathrm {th}}$ was calculated using the Constant-Current Method \cite{bucher2020generalized}.



\begin{table}
\centering
\renewcommand{\arraystretch}{1.5} 
\caption{Device Parameters Used in the TCAD Simulation}
\label{table 1 Device Parameters Used in the TCAD Simulation}
\begin{tabularx}{\columnwidth}{|>{\centering\arraybackslash}X|>{\centering\arraybackslash}X|}
\hline
\textbf{Parameters} & \textbf{Values} \\
\hline
$l_g (\text{nm})$ & $[12, 14, 16, 18, 20, 22]$ \\
\hline
$r (\text{nm})$ & $[2, 3, 4, 5]$ \\
\hline
$h (\text{nm})$ & $[2, 3, 4]$ \\
\hline
$w (\text{nm})$ & $[2, 3, 4]$ \\
\hline
$t_{ox} (\text{nm})$ & $[0.5, 1, 1.5]$ \\
\hline
$N_{sd} (\times 10^{20}\, \text{cm}^{-3})$ & $[0.5, 1, 2]$ \\
\hline
$\varepsilon_{ox}$ & $[3.9, 7.5, 22]$ \\
\hline
$V_{\mathrm ds}(\text{V})$ & $[0, 0.05, 0.1, 0.15, \cdots, 0.7]$ \\
\hline
$V_{\mathrm gs}(\text{V})$ & $[0, 0.05, 0.1, 0.15, \cdots, 0.7]$ \\
\hline
\end{tabularx}
\end{table}

\begin{table*}[!ht]
\centering
\caption{Comparison of Different Models}
\label{table 2 Comparison of Different Models}
\renewcommand{\arraystretch}{2} 
\resizebox{\textwidth}{!}{
\begin{tabular}{c|ccccc|ccccc|ccccc}
\hline
\multirow{2}{*}{Model} & \multicolumn{5}{c|}{Triangle} & \multicolumn{5}{c|}{Rectangle} & \multicolumn{5}{c}{Circle} \\ \cline{2-16} 
                       & $I_{\mathrm{on}}$ & $I_{\mathrm{off}}$ & $\min_{e}$ & $\max_{e}$ & MRE & $I_{\mathrm{on}}$ & $I_{\mathrm{off}}$ & $\min_{e}$ & $\max_{e}$ & MRE & $I_{\mathrm{on}}$ & $I_{\mathrm{off}}$ & $\min_{e}$ & $\max_{e}$ & MRE \\ \hline
MNN                    & 5.20\%   & 5.92\%    &    4.33\%    &    5.39\%    & 4.66\% 
                       & 2.06\%  & 6.27\%   &   2.45\%     &  2.82\%      & 2.66\% 
                       & 2.63\%   & 7.65\%    &      3.16\%  &   3.44\%     & 3.32\% \\ 
DrPCE-Net-3            & 5.46\%   & 7.74\%    &    4.50\%  &      4.88\%  & 4.70\% 
                       & 2.40\%   & 7.10\%   &     2.47\%   &     2.74\%   & 2.64\% 
                       & 3.13\%   & 7.51\%    &    2.71\%    &     3.15\%   & 2.99\% \\ 
DrPCE-Net-5            & 5.58\%   & 6.21\%    &      4.71\%   &     5.32\%   & 4.93\% 
                       & 2.44\%  & 5.44\%  &      2.26\%  &     2.41\%   & 2.33\%
                       & 3.05\%   & 6.57\%    &    2.96\%    &      2.81\%  & 2.88\% \\ 
PPC-Net               & 4.64\%   & 5.30\%    &     3.74\%   &    4.23\%    & 4.01\% 
                       &     1.86\%     &     5.10\%       &  1.66\%      &    1.84\%    &   1.74\%    
                       &     2.42\%     &     5.36\%      &   2.54\%     &     2.96\%   &    2.71\%    \\ 
PRIME                    & \textbf{3.00\%}   & \textbf{3.94\%}    &   \textbf{2.68\%}     &  \textbf{2.85\%}    & \textbf{2.76\%} 
                       &    \textbf{1.34\%}      &    \textbf{4.27\%}       &  \textbf{ 1.51\%}   &   \textbf{1.68\%}     &      \textbf{1.58\%}
                       &     \textbf{1.72\%}     &     \textbf{4.11\%}      &    \textbf{1.67\%}    &    \textbf{1.93\%}    &     \textbf{1.80\%}  \\ \hline
\end{tabular}}
\end{table*}

\begin{figure*}[!ht]
\centering
\includegraphics[width=\textwidth]{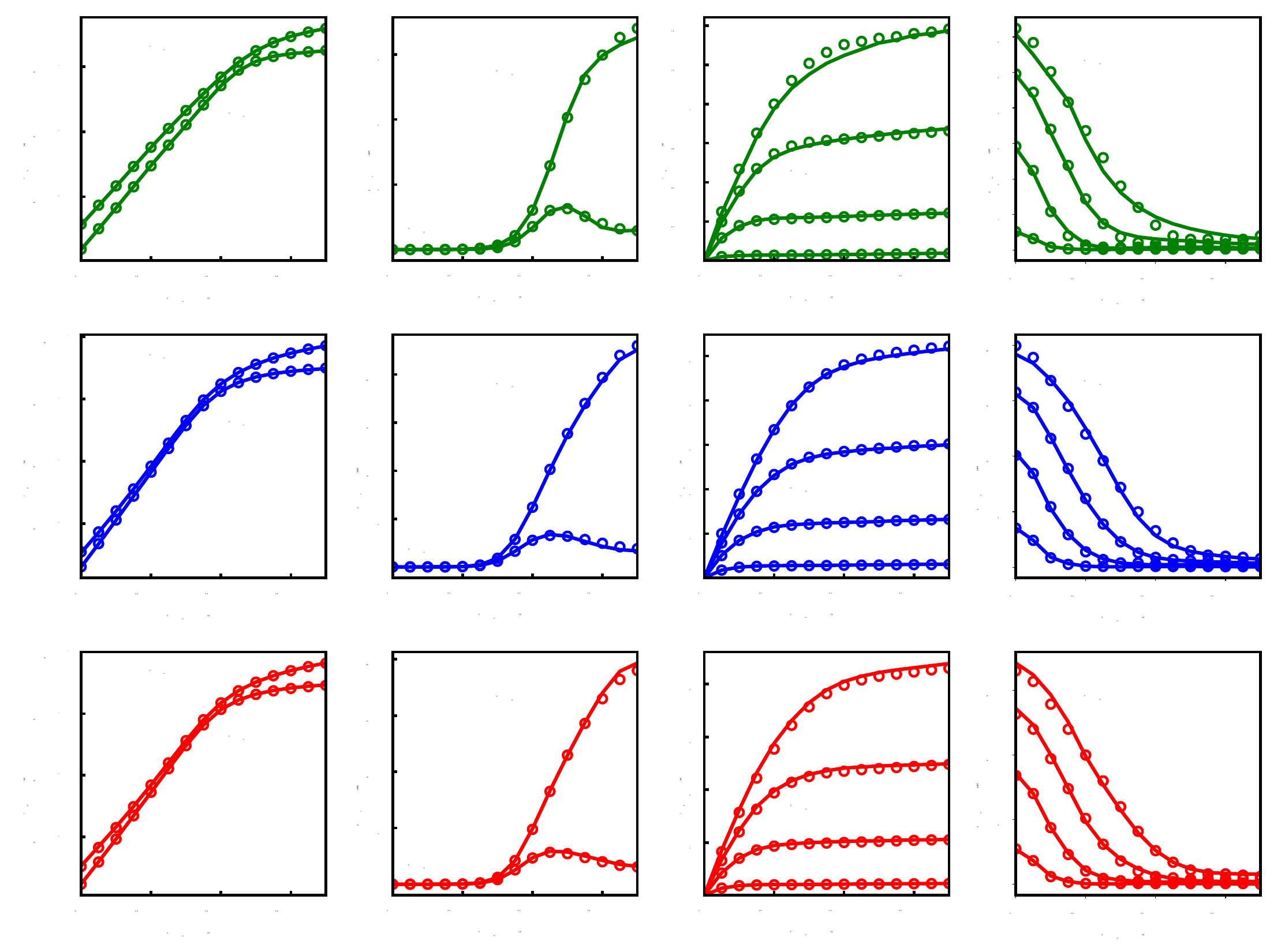}
\caption{The predicted current characteristics of GAA transistors with triangular, rectangular, and circular cross sections by the PRIME model. The hollow circles represent TCAD data, while the solid lines represent the prediction results of the PRIME model.} 
\label{fig3: MoE Predicted}
\end{figure*}



\subsection{Model Parameters and Evaluation Metrics}

To evaluate the predictive accuracy of PRIME, we designed a benchmark experiment and compare the PRIME model with two state-of-the-art baseline models: the multigradient neural network (MNN) proposed by Yang et al. \cite{yang2021transistor} and the differential residual polynomial chaos expansion network (DrPCE-Net) proposed by Wang et al. \cite{wang2023drpce}. The first model is a general transistor model based on MNN, incorporating higher-order derivative information during network training. The second model performs an initial fit using generalized polynomial chaos \cite{xiu2003modeling} and then applies a residual neural network \cite{he2016deep} for correction and fine-tuning.

In our model, PRIME consists of three expert models and a gating network. Each expert model shares the same architecture, featuring two hidden layers with 20 neurons per layer. The gating network is built as a three-layer MLP, with each hidden layer also containing 20 neurons. PPC-Net uses the same architecture as PRIME and is also employed for comparative experiments.


The baseline MNN model consists of four hidden layers, with neuron counts of 16, 32, 32 and 16 in each layer, respectively. For DrPCE-Net, we selected two models based on different polynomial orders: one uses the third-order polynomial as presented in earlier works~\cite{wang2023drpce}, where the residual neural network consists of four layers, including two hidden layers. It is labeled as DrPCE-Net-3 in subsequent presentation. 
The other DRPCE-Net uses a fifth-order polynomial, where the residual neural network consists of five layers, including three hidden layers, each with 20 neurons. It is labeled as DrPCE-Net-5. It should be noted that the parameter settings of DrPCE-Net-3 are consistent with those reported in \cite{wang2023drpce}, while the other models have a comparable number of parameters to the PRIME model. All networks are optimized using the Adam optimizer \cite{kingma2014adam} with a learning rate of $10^{-3}$. All experiments were conducted on a Linux server with AMD EPYC 7402 24-core processor and one Nvidia GeForce RTX 3080 Ti GPU.

Following the same evaluation procedures as in earlier works~\cite{yang2021transistor,wang2023drpce}, we conducted experiments on GAA devices with three different cross-sectional shapes. Given the varying convergence rates of different models across the three datasets, we set the number of optimization steps to 5,000 to ensure each model converges to its optimal state at the training stage. To reduce training variance due to random seeds, all experiments were repeated five times with different random seeds.

\subsection{Performance Evaluation}

Fig.~\ref{fig3: MoE Predicted} shows the current characteristics of GAA transistors with three different cross-sectional shapes, as predicted by the PRIME model. Here $g_{m}=\partial I_{\mathrm {ds}} / \partial V_{\mathrm {gs}}$ and $g_{\mathrm{ds}}=\partial I_{\mathrm {ds}} / \partial V_{\mathrm {gs}}$. From those figures, one can see PRIME accurately predicts the increase in drain current with the bias voltage under different cross-sectional shapes. Its prediction on the derivatives also aligns with that obtained from the finite difference method~\cite{thomas2013numerical}.


Table \ref{table 2 Comparison of Different Models} presents the prediction errors of several key parameters for all models. 
The mean relative error (MRE) is defined as the statistical average of five random experiments: $1/n \sum_{i=1}^{n} |(y-\tilde{y})/ y|$, where $\min_{e}$ indicates the minimum error from the five experiments and $\max_{e}$ indicates the maximum error. \( I_{\mathrm{on}} \) is the current at \( V_{\mathrm {ds}}=0.7\,\text{V} \) and \( V_{\mathrm {gs}}=0.7\,\text{V} \), while \( I_{\mathrm{off}} \) is the current at \( V_{\mathrm {ds}}=0.7\,\text{V} \) and \( V_{\mathrm {gs}}=0\,\text{V} \). Together they represent the mean relative error from the five experiments.

As shown in Table \ref{table 2 Comparison of Different Models}, the PRIME model exhibits the lowest prediction errors among all comparison models, indicating significant performance advantages in GAA transistors with different cross-sectional shapes. In terms of relative mean error, for the triangular cross-section, PRIME reduces prediction error by 69\% compared to the MNN model, and by 70\% and 78\% compared to DrPCE-Net-3 and DrPCE-Net-5, respectively. When compared to PPC-Net, PRIME improves prediction accuracy by 45\%. It also shows similar improvements for rectangular and circular cross-sections, with error reductions of 68\% and 84\% compared to MNN, and over 60\% compared to DrPCE-Net-3 and DrPCE-Net-5. Those results validate the robustness and excellent generalization capability of the PRIME model under various scenarios, thus highlights its potential for practical applications.

\subsection{Physical Applications}

We also assess the performance of PRIME in circuit simulation to explore its potential as an alternative to current compact models. We carried out SPICE simulations (Simulation Program with Integrated Circuit Emphasis) on an inverter (INV) and SRAM. 

\begin{figure}[htb]
\centering
\includegraphics[width=0.68\columnwidth]{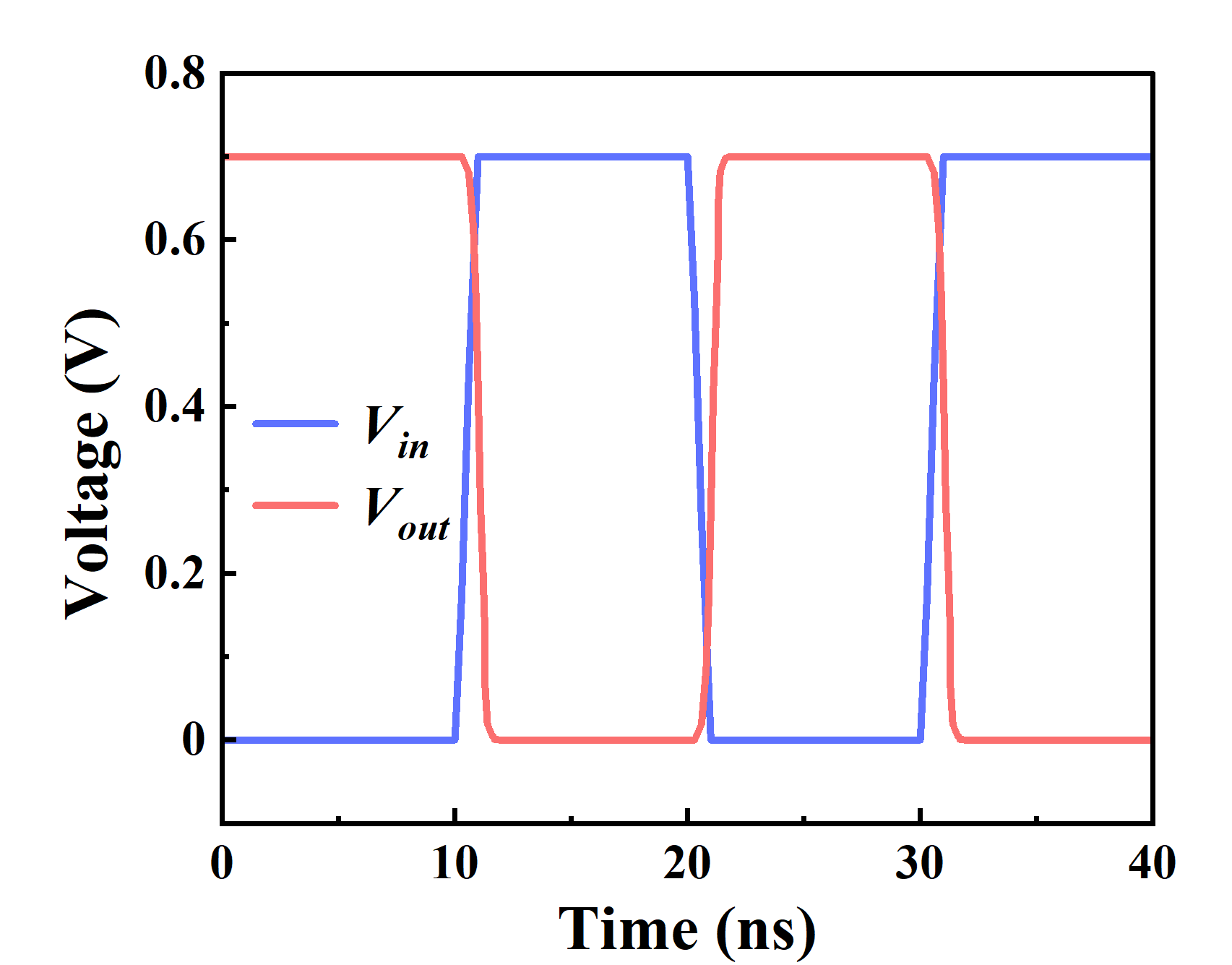}
\caption{Inverter simulation results.}
\label{fig1: INV}
\end{figure}

\begin{figure}[htb]
\centering
\includegraphics[width=\columnwidth]{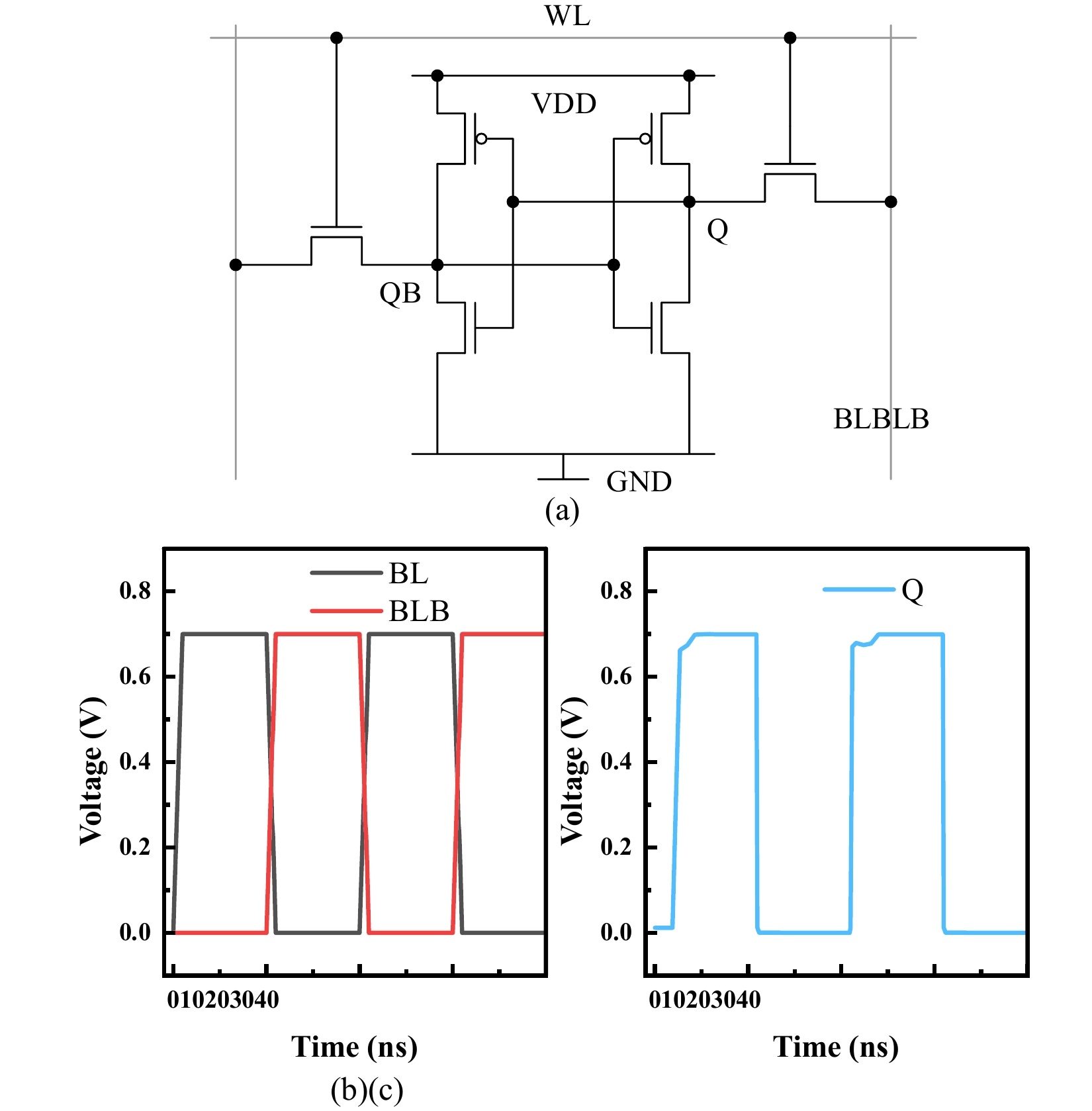}
\caption{(a) SRAM Structure. (b) Input of write operation simulation. (c) Results of write operation simulation.}
\label{fig1: SRAM}
\end{figure}

As shown in Fig.~\ref{fig1: INV}, the inverter is simulated with a 0.9 $\text{fF}$ added at the output terminal to facilitate the charging and discharging process. The results validate the expected relationship $V_{\text{in}}$ = $\overline{V_{\text{out}}}$. As shown in Fig.~\ref{fig1: SRAM}, the SRAM write operation is simulated with a 0.9 $\text{fF}$ load added at the output terminal and the results confirm the expected relationship  Q = BL, which means the "1" bit was successfully written to the SRAM. Overall, those results using the PRIME predictions provides decent characterization of GAA $I-V$ response for SPICE simulation.

\section{CONCLUSION} \label{section 5}

This study introduces PRIME, a physics-driven intelligent mixture of experts model, which focuses on enhancing transistor characteristics prediction accuracy across different operating regions. By leveraging a dynamic weighting mechanism in the gating network, PRIME adaptively activates suitable expert models based on distinct input data features, thus captures complex regional characteristics. This approach addresses limitations often associated with traditional neural networks, such as interpretability and generalization. Compared to conventional neural network methods, PRIME achieves significant accuracy improvements. Experimental results demonstrate that PRIME provides high-precision current predictions for GAA transistors with various cross-sectional shapes. Its robustness and versatility in cross-region characteristics modeling offer a new and efficient modeling solution for circuit simulation.

\newpage

\bibliographystyle{IEEEtran}
\bibliography{ref}

\end{document}